\documentclass[letterpaper, 10pt, conference]{ieeeconf}

\usepackage{amsmath}
\usepackage{amssymb}
\usepackage{amsfonts}
\usepackage{flushend}
\usepackage{mathrsfs}
\usepackage{graphicx}
\usepackage{subfigure}
\usepackage{setspace}
\usepackage{multirow}
\usepackage{footnote}
\usepackage{algpseudocode,algorithm}
\usepackage{setspace}
\usepackage{multirow}
\usepackage{verbatim}

\graphicspath{{figures/}}

\usepackage{theorem}

 \usepackage[dvips]{color}
 \newcommand{\black}{\color{black}}
 \newcommand{\blue}{\color{blue}}

\renewcommand{\thesection}{\arabic{section}}
\renewcommand{\baselinestretch}{1.0}
\IEEEoverridecommandlockouts                              % This command is only
\begin{document}
\title{\small \blue For presentation in \emph{CDC 2017} \black \\
\LARGE \textbf{Data-driven root-cause analysis for distributed system anomalies} }
\author{Chao Liu, Kin Gwn Lore, and Soumik Sarkar
\thanks{Department of Mechanical Engineering, Iowa State University, Ames, IA 50011, USA. {\small chaoliu08@gmail.com, lorekingwn@gmail.com, soumiks@iastate.edu}}
}

\maketitle

\begin{abstract}
  Modern distributed cyber-physical systems encounter a large variety of anomalies and in many cases, they are vulnerable to catastrophic fault propagation scenarios due to strong connectivity among the sub-systems. In this regard, root-cause analysis becomes highly intractable due to complex fault propagation mechanisms in combination with diverse operating modes. This paper presents a new data-driven framework for root-cause analysis for addressing such issues. The framework is based on a spatiotemporal feature extraction scheme for distributed cyber-physical systems built on the concept of symbolic dynamics for discovering and representing causal interactions among subsystems of a complex system. We present two approaches for root-cause analysis, namely the sequential state switching ($S^3$, based on free energy concept of a Restricted Boltzmann Machine, RBM) and artificial anomaly association ($A^3$, a multi-class classification framework using deep neural networks, DNN). Synthetic data from cases with failed pattern(s) and anomalous node are simulated to validate the proposed approaches, then compared with the performance of vector autoregressive (VAR) model-based root-cause analysis. Real dataset based on Tennessee Eastman process (TEP) is also used for validation. The results show that: (1) $S^3$ and $A^3$ approaches can obtain high accuracy in root-cause analysis and successfully handle multiple nominal operation modes, and (2) the proposed tool-chain is shown to be scalable while maintaining high accuracy.
\end{abstract}

\section{Introduction}%\vspace{-8pt}
With the advent of ubiquitous sensing, advanced computation and strong connectivity, modern distributed cyber-physical systems (CPSs) such as power plants, integrated buildings, transportation networks and power-grids have shown tremendous potential of increased efficiency, robustness and resilience. From the perspective of performance monitoring, anomaly detection and root-cause analysis of such systems, technical challenges arise from a large number of subsystems that are highly interactive and operate in diverse modes.

For the purpose of root-cause analysis in distributed complex systems, Granger causality~\cite{barnett2009granger} is applied to model the system-wide behavior and capture the variation that can be used to implement root-cause analysis. With multivariate time series data, studies show that the causality from and to the fault variable presents differences and can be used to reason the root-cause~\cite{yuan2014root,mori2014identification,landman2014fault}. For anomaly detection in time series, Qiu et. al.~\cite{qiu2012granger} derived neighborhood similarity and coefficient similarity from Granger-Lasso algorithm, to compute anomaly score and ascertain threshold for anomaly detection. A causality analysis index based on dynamic time warping is proposed by Li et. al.~\cite{li2015dynamic} to determine the causal direction between pairs of faulty variables in order to overcome the shortcoming of Granger causality in nonstationary time series. The proposed approaches provide efficient tools in discovering causality in complex systems, while an approach in inferencing (interpreting the variation in causality into decisions on failed patterns of fault variable/node) is less investigated.

In this context, we present a semi-supervised tool for root-cause analysis in distributed CPSs based on a data driven framework proposed for system-wide anomaly detection in distributed complex system~\cite{liu2017bridge}, and using a spatiotemporal feature extraction scheme built on the concept of symbolic dynamics for discovering and representing causal interactions between the subsystems. The proposed tool aims to (i) capture multiple operating modes as nominal in complex CPSs, (ii) only use nominal data and artificially generated fault data to train the model without requiring true labeled anomalous data, and (iii) implement root-cause analysis in a semi-supervised way in a diversity of fault types (e.g., single fault and multiple faults). We present two approaches for root-cause analysis, namely the \textit{sequential state switching} ($S^3$, based on free energy concept of a Restricted Boltzmann Machine, RBM~\cite{HS06}) and \textit{artificial anomaly association} ($A^3$, a multi-class classification framework using deep neural networks, DNN).

\section{Background and preliminaries}%\vspace{-8pt}
\label{secSDF}
\subsection{Spatiotemporal pattern network (STPN)}%\vspace{-5pt}
\label{secSTPN}
STPN modeling involves learning sequential characteristics in a univariate and pairwise manner from a multi-dimensional time-series data. While details can be found in~\cite{sarkar2014sensor,LGJS16conf}, we are providing a brief description here for completeness.

Consider a multivariate time series, $\mathbf{X}=\{X^{\mathbb{A}}(t)$, $t \in \mathbb{N}$, $\mathbb{A}=1,2,\cdots, n\}$, where $n$ is the number of variables or dimension of the time series, corresponding to the number of nodes in graphical modeling. Let $\mathbb{X}$ denote a set of partitioning/discretization functions~\cite{RRSY09}, $\mathbb{X}: \ X(t)\to S$, that transform a general dynamic system (time series $X(t)$) into a symbol sequence $S$ with an alphabet set $\Sigma$. Various partitioning approaches have been proposed in the literature, such as uniform partitioning (UP), maximum entropy partitioning (MEP, used for the present study), maximally bijective discretization (MBD) and statistically similar discretization (SSD)~\cite{SS16}.
Subsequently, a probabilistic finite state automaton (PFSA) is defined to describe states (representing various parts of the data space) and probabilistic transitions among them (can be learnt from data) via $D$-Markov machine and $xD$-Markov machine. Related definitions and further details on learning schemes can be found in~\cite{sarkar2014sensor}.
With this setup, an STPN is defined as:

\textbf{Definition}. A PFSA based STPN is a 4-tuple $W_{D} \equiv ( Q^{a}, \Sigma^{b}, \Pi^{ab}, \Lambda^{ab})$: (a, b denote nodes of the STPN)
\begin{enumerate} %\vspace{-6pt}
  \item $Q^{a}=\{q_{1}, q_{2}, \cdots, q_{|Q^{a}|}\}$ is the state set corresponding to symbol sequences ${S^{a}}$~\cite{sarkar2014sensor}.  %\vspace{-3pt}
  \item $\Sigma^{b}=\{\sigma_{0}, \cdots, \sigma_{|\Sigma^{b}|-1}\}$ is the alphabet set of symbol sequence ${S^{b}}$. %\vspace{-3pt}
  \item $\Pi^{ab}$ is the symbol generation matrix of size $|Q^{a}| \times |\Sigma^{b}|$, the $ij^{th}$ element of $\Pi^{ab}$ denotes the probability of finding the symbol $\sigma_{j}$ in the symbol string  ${s^{b}}$ while making a transition from the state $q_{i}$ in the symbol sequence ${S^{a}}$; while self-symbol generation matrices are called atomic patterns (APs) i.e., when $a=b$, cross-symbol generation matrices are called relational patterns (RPs) i.e., when $a \neq b$.  %\vspace{-3pt}
  \item $\Lambda^{ab}$ denotes a metric that can represent the importance of the learnt pattern (or degree of causality) for $a \rightarrow b$ which is a function of $\Pi^{ab}$.
\end{enumerate}%\vspace{-8pt}

An example of the STPN model is shown in ~\cite{LGJS16conf}. %refer to CPS journal in the future

\subsection{Unsupervised anomaly detection with spatiotemporal causal graphical modeling}%\vspace{-5pt}
A data-driven framework for system-wide anomaly detection is proposed in~\cite{LGJS16conf}, noted as the \textbf{STPN+RBM} model, including the following steps:
\begin{enumerate}%\vspace{-6pt}
\item Learn APs and RPs (individual node behaviors and pair-wise interaction behaviors) from the multivariate training symbol sequences. %\vspace{-3pt}
\item Consider short symbol sub-sequences from the training sequences and evaluate $\Lambda^{ij}\ \forall i,j$ for each short sub-sequence. %\vspace{-3pt}
\item For one sub-sequence, based on a user-defined threshold on $\Lambda^{ij}$, assign state $0$ or $1$ for each AP and RP; thus every sub-sequence leads to a binary vector of length $L$, where $L = \#AP + \#RP$. %\vspace{-3pt}
\item An RBM is used for modeling system-wide behavior with nodes in the visible layer corresponding to APs and RPs. %\vspace{-3pt}
\item The RBM is trained using binary vectors generated from nominal training sub-sequences. %\vspace{-3pt}
\item Online anomaly detection is implemented by computing the probability of occurrence of a test STPN pattern vector via trained RBM.
\end{enumerate}%\vspace{-6pt}

The anomaly detection process and details can be found in~\cite{LGJS16conf}.

\section{Methods}%\vspace{-10pt}
\label{secMethod}
\subsection{Root-cause analysis problem formulation}
With the definition of STPN in Section \ref{secSTPN}, an inference based metric is employed for evaluating the patterns (APs \& RPs) \cite{LGJS16conf}. The inference based metric computation includes a modeling phase and an inference phase. In the modeling phase, the time-series in the nominal condition is applied, noted as $X=\{X^{\mathbb{A}}(t)$, $t \in \mathbb{N}$, $\mathbb{A}=1,2,\cdots, f \}$, where $f$ is the number of variables or the dimension of the time series. The time series is then symbolized into $S=\{S^{\mathbb{A}}\}$ and then state sequences are generated with the STPN formulation, noted by $Q=\{Q^a,\ a=1,2,\cdots, f \}$. In the learning phase, short time-series is considered, $\tilde{X}=\{\tilde{X}^{\mathbb{A}}(t)$ for the short time-series in nominal condition, $t \in \mathbb{N}^{*}$, $\mathbb{A}=1,2,\cdots, f \}$, where $\mathbb{N}^{*}$ is a subset of $\mathbb{N}$, and $\hat{X}=\{\hat{X}^{\mathbb{A}}(t)$ for the short time-series in anomalous condition. The corresponding short symbolic subsequences is noted as $\tilde{S}=\{\tilde{S}^{\mathbb{A}}\}$ and $\hat{S}=\{\hat{S}^{\mathbb{A}}\}$ for nominal condition and anomalous condition respectively, and the state sequences $\tilde{Q}$ and $\bar{Q}$ correspondingly. An importance metric $\Lambda^{ab}$ is defined, which suggests the importance of the pattern $\Pi^{ab}$ or the degree of causality in $a \rightarrow b$ as evidenced by the short subsequence.

The inference based metric $\Lambda^{ab}(\tilde{Q}, \tilde{S})$ for a pattern $a\to b$ can be obtained as follows \cite{LGJS16conf},
\begin{align}
\begin{split}
& \Lambda^{ab}(\tilde{Q}, \tilde{S})  =\\
  &   K \prod_{m=1}^{|Q^{a}|}{\frac{(\tilde{N}^{ab}_{m})!(N^{ab}_{m}+|\Sigma^{b}|-1)!}{(\tilde{N}^{ab}_{m}+N^{ab}_{m}+|\Sigma^{b}|-1)!}}
                      \prod_{n=1}^{|\Sigma^{b}|}{\frac{(\tilde{N}^{ab}_{mn}+N^{ab}_{mn})!}{(\tilde{N}^{ab}_{mn})!(N^{ab}_{mn})!}}
\label{equProbOnline1}
\end{split}
\end{align}
where, $K$ is a proportional constant, $N_{mn}^{ab} \triangleq |\{(Q^{a}(k),S^{b}(k+1): S^{b}(k+1) = \sigma^{b}_n\ | \ Q^{a}(k) = q^a_m \}|$, $N_{m}^{ab}=\sum_{n=1}^{|\Sigma^{b}|}(N_{mn}^{ab})$, $\tilde{N}^{ab}_{mn}$ and $\tilde{N}^{ab}_{m}$ are similar to $N_{mn}^{ab}$ and $N_{m}^{ab}$, $|Q^{a}|$ is number of states in state sequence $\tilde{Q}$, and $|\Sigma^{b}|$ is number of symbols in symbol sequence $\tilde{S}$.

For an anomalous condition, the inference based metric $\hat{\Lambda}^{ab}(\hat{Q}, \hat{S})$ for a pattern $a\to b$ is obtained as follows,
\begin{align}
\begin{split}
& \hat{\Lambda}^{ab}(\hat{Q}, \hat{S})  =\\
  &   K \prod_{m=1}^{|Q^{a}|}{\frac{(\hat{N}^{ab}_{m})!(N^{ab}_{m}+|\Sigma^{b}|-1)!}{(\hat{N}^{ab}_{m}+N^{ab}_{m}+|\Sigma^{b}|-1)!}}
                      \prod_{n=1}^{|\Sigma^{b}|}{\frac{(\hat{N}^{ab}_{mn}+N^{ab}_{mn})!}{(\hat{N}^{ab}_{mn})!(N^{ab}_{mn})!}}
\label{equProbOnline1}
\end{split}
\end{align}
where $\hat{N}$ is with the similar definition in Eq. \ref{equProbOnline1}, while it is emanated from the time series $\hat{X}=\{\hat{X}^{\mathbb{A}}(t)$, $t \in \mathbb{N}^{*}$, $\mathbb{A}=1,2,\cdots, f \}$ in the anomalous condition.

Let $\delta \Big(\ln( \Lambda ^{ab}) \Big)$ denote the variation of the metric, $\delta \Big( \ln( \Lambda ^{ab}) \Big)= \ln\Big(\Lambda^{ab}(\tilde{Q}, \tilde{S})\Big) - \ln\Big(\hat{\Lambda}^{ab}(\hat{Q}, \hat{S})\Big)$. We also define the set of all metrics in the nominal condition as $\Lambda = \{\Lambda^{ab}\} \ \forall a, b$ and the set of all metrics in the anomalous condition as $\hat{\Lambda} = \{\hat{\Lambda}^{ab}\} \ \forall a, b$.

The main idea behind the proposed root-cause analysis algorithm is to perturb the space of test (anomalous) patterns in an artificial manner to bring it close to the space of nominal patterns. During this process, we aim to identify the nodes/patterns involved in successful perturbations that bring the test pattern space sufficiently close to the space of nominal patterns. Then we label those identified nodes/patterns as the possible root-cause(s) for the detected anomaly. More formally, let us suppose that an inference metric $\hat{\Lambda}^{ab}$ changes to $\hat{\Lambda}^{ab\prime}$ due to an artificial perturbation in the pattern $a\to b$. We now consider a subset of patterns, for which we have the set of inference metrics $\{\hat{\Lambda}^{ab}\} \subset \hat{\Lambda}$. Let a perturbation in this subset changes the overall set of metrics to $\hat{\Lambda}^{\prime}$. Therefore, we have the following:
\begin{equation}
%\nonumber
\hat{\Lambda}^{\prime}=
    \begin{cases}
      \hat{\Lambda}^{ab} \ \ \text{if}\ \  \hat{\Lambda}^{ab} \not\in \{\hat{\Lambda}^{ab}\}\\
      \hat{\Lambda}^{ab\prime} \ \ \text{if}\ \  \hat{\Lambda}^{ab} \in \{\hat{\Lambda}^{ab}\}
    \end{cases}
\end{equation}

The root cause analysis is then formulated as a minimization problem between the set of nominal inference metrics $\Lambda$ and the set of perturbed inference metrics $\hat{\Lambda}^{\prime}$ which can be expressed as follows:
\begin{align}
\begin{split}
%&\min_{\mathscr{D}}
\{\hat{\Lambda}^{ab}\}^{\star} = \min_{\{\hat{\Lambda}^{ab}\}} \mathscr{D}\Big(\hat{\Lambda}^{\prime},\Lambda\Big),%\ \hat{\Lambda}^{ab} \in \{\hat{\Lambda}^{ab}\}.
\label{equInference}
\end{split}
\end{align}
where $\mathscr{D}$ is a distance metric (e.g., Kullback-Leibler Distance--KLD\cite{kullback1951information,frey2005comparison}) to estimate the difference between $\Lambda$ and $\hat{\Lambda}^{\prime}$. The nodes (e.g., $a$ or $b$) or patterns (e.g., $a\to b$) involved in the optimal subset $\{\hat{\Lambda}^{ab}\}^{\star}$ will be identified as possible root cause(s) for the detected anomaly. However, solving this optimization problem in an exact manner may not be computationally tractable for large systems. Therefore, we propose two approximate algorithms: the sequential state switching ($S^3$) - a sequential suboptimal search method and artificial anomaly association ($A^3$) - a semi-supervised learning based method.

\subsection{Sequential state switching ($S^3$)}
In the STPN+RBM framework described above, anomaly manifests itself as a low probability (high energy) event. Therefore, the idea for $S^3$ is to find potential pattern(s) that, if changed, can transition the system from a high to a low energy state. The probabilities of AP and RP's existence are discovered by the STPN, and an anomaly will influence the causality of specific patterns (e.g., in STPN, the probability of the pattern might be switched/flipped from 0 to 1). Hence, by switching/flipping a pattern, its contribution on the energy states of the system can be identified and a large contribution may indicate the root-cause of an anomaly.

For an $n$-node graphical model, all the APs and RPs together form a binary vector $v$ of length $L = n^2$ ($L=\#AP+\#RP$, where $\#AP=n$, $\#RP=n \times (n-1)$). One such binary vector is treated as one training example for the system-wide RBM (with $n^2$ number of visible units) and many such examples are generated from different short sub-sequences extracted from the overall training sequence. Then, the RBM is trained by maximizing the maximum likelihood of the data.

With the weights and biases of RBM, free energy can be computed with the following expression~\cite{hinton2012practical}:
\begin{equation}
%\nonumber
F(v)=-\sum_{i}{v_{i}a_{i}}-\sum_{j}\log(1+e^{b_{j}+\sum_{i}{v_{i}w_{ij}}})
\label{equFreEngy}
\end{equation}
The free energy in nominal conditions is noted as $\tilde{F}$. In anomalous conditions, a failed pattern will shift the energy from a lower state to a higher state. Assume that the patterns can be categorized into two sets, $\textbf{v}^{nom}$ and $\textbf{v}^{ano}$. By flipping the set of anomalous patterns $\textbf{v}^{ano}$, a new expression for free energy is obtained:
\begin{align}
%\nonumber
\begin{split}
F^{s}(v)=&-\sum_{g}{v_{g}a_{g}}-\sum_{j}\log(1+e^{b_{j}+\sum_{g}{v_{g}w_{gj}}})\\
           &-\sum_{h}{v^{\star}_{h}a_{h}}-\sum_{j}\log(1+e^{b_{j}+\sum_{h}{v^{\star}_{h}w_{hj}}}), \\ &\quad \{v_{g}\} \in v^{nom}, \{v^{\star}_{h}\} \in  v^{\star,ano}
\label{equFreEngyAno}
\end{split}
\end{align}
Here, $v^{\star}$ has the opposite state to $v$ and represents that the probability of the pattern has been significantly changed. In this work, the probabilities of the patterns are binary (i.e. 0 or 1). Hence, we have that $v^{\star}=1-v$. The sequential state switching is formulated by finding a set of patterns $v^{ano}$ via $\min(F^{s}(v^{ano},v^{nom})-\tilde{F})$. With bijection function between input units of RBM ($\mathbf{v}$) and patterns in STPN ($\mathbf{\Lambda}$, $\Lambda^{a,a}$ for AP and $\Lambda^{a,b}$ for RP), failed patterns in STPN can be inferred.

\subsection{Artificial anomaly association ($A^3$)}%\vspace{-5pt}
In this proposed approach, we frame the root-cause analysis problem as a supervised classification problem~\cite{lore2015hierarchical}. The crux of the idea is that anomalies are artificially injected (by introducing anomalous patterns) into the learnt nominal model and label it with that particular anomaly. Then we employ a supervised learning method to map the entire space of nominal-anomalous patterns to the space of labels. Upon learning, this map can be used to detect the root cause (as the labels) given a test nominal-anomalous pattern. Since all possible anomalous data is mostly infeasible to obtain for a real system, the proposed synthetic scheme provides a useful alternative. Regarding the choice of supervised learning scheme, we note that realistic physical systems may have multiple nominal nodes and it will most certainly be arduous to extract important features from the nominal-anomalous patterns to obtain sufficiently high accuracy. In this context, we turn our attention to deep learning methods.

Artificial anomaly association is based on a method proposed in~\cite{lore2015hierarchical} to solve a multi-label classification problem using convolutional neural networks (CNN). Instead of inferring a single class from the trained model, the framework solves $n_{out}$ classification sub-problems if an output vector of length $n_{out}$ is required using the previously learned model. The implementation of this formulation requires only a slight modification in the loss function: for an output sequence with length $n_{out}$, the loss function to be minimized for a data set $\mathcal{D}$ is the negative log-likelihood defined as:
\begin{equation}
\begin{split}
& \ell_{total}(\theta=\{W,b\},\mathcal{D}) = \\
& - \sum_{j=1}^{n_p} \sum_{i=0}^{\mathcal{|D|}} \left[ \log{\left(P(Y=y^{(i)}|x^{(i)},W,b)\right)} \right]_j
\end{split}
\end{equation}

where $\mathcal{D}$ denotes the training set, $\theta$ is the model parameters with $W$ as the weights and $b$ for the biases. $y$ is predicted target vector whereas $x$ is the provided input pattern. The total loss is computed by summing the individual losses for each sub-label.

The input is presented as an $n^2$-element vector with values of either 0 or 1 which denotes whether a specific pattern is activated. We desire to map the input vector to an output vector of the same length (termed as the \textit{indicator label}), where the value of each element within the output vector indicates whether a specific pattern is anomalous. For nominal modes, the input vector may be comprised of different combinations of 0's and 1's, and the indicator labels will be a vector of all 1's (where the value 1 denotes no anomaly). However, if a particular element $i$ within the input vector gets flipped, then the indicator label corresponding to the $i$-th position in the output vector will be flipped and switches from 1 (normal) to 0 (anomalous). In this way, we can identify that the $i$-th pattern is anomalous. With this setup, a classification sub-problem (i.e. is this pattern normal, or anomalous?) can be solved for each element in the output vector given the input data.

\begin{figure*}[htbp]
\centering
\includegraphics[width=0.75\textwidth,trim={0 20 0 10}]{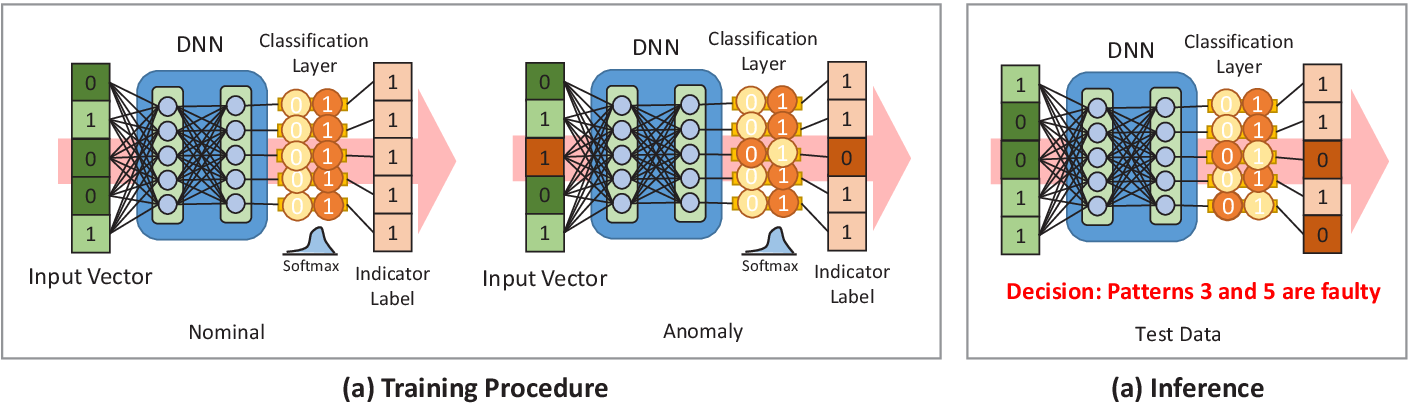}  \caption{Framing the problem as an artificial anomaly association ($A^3$) problem. (a) When training the model, the value of an element in both the input vector and its corresponding indicator label is randomly flipped to simulate anomaly. (b) When inferencing, a test input is fed into the DNN and a classification sub-problem is solved to obtain the indicator vector. Values of 0 in the output vector traces back to the exact patterns that are faulty.} %\vspace{-8pt}
\label{fig:fig_smcformulation}
\end{figure*}%\vspace{-10pt}

\section{Results and discussions}%\vspace{-10pt}
\label{secResults}
Synthetic datasets are generated with vector autoregressive (VAR) process to simulate anomaly in pattern(s)/node(s) for performance evaluation of $S^3$ and $A^3$ methods. The inference with VAR is then used for comparison. A dataset based on TEP is also used in validating the proposed methods.

\subsection{Anomaly in pattern(s)}%\vspace{-5pt}
\label{subsecAnoPat}

\textbf{Dataset:} Anomaly in pattern(s) is defined as the change of one or more causal relationship, while defining anomaly, this translates to a changed/switched pattern in the context of STPN. A 5-node system is defined including six different nominal modes. Anomalies are simulated by breaking specific patterns in the graph; 30 cases are formed including 5 cases in one failed pattern, 10 cases in two failed patterns, 10 cases in three failed patterns, and 5 cases in four failed patterns. Multivariate time series data (denoted as \texttt{dataset1}) are generated using VAR process that follows the causality definition in the graphical models.

\textbf{Performance Evaluation:} Root-cause identification performances of $S^3$ and $A^3$ methods are evaluated using \texttt{dataset1}. Recall, precision and F-measure are evaluated with the definitions in ~\cite{fawcett2006introduction}.

\setlength\tabcolsep{3pt}
\begin{table}[htbp]%\vspace{-10pt}
\caption{Root-cause analysis results in $S^3$ method and $A^3$ method.}%\vspace{3pt}
\centering
\label{tabAccu}
\begin{tabular}{c c c c c c c}
\hline
  Method &Training/Testing & Recall & Precision &F-measure\\
  \hline
  $S^3$ &11,400/57,000 &99.40\% &97.10\% &98.24\%\\ 
  $A^3$ &296,400/57,000 &90.46\% &95.95\% &93.12\%\\
\hline
\end{tabular}%\vspace{-5pt}
\end{table}%\vspace{-10pt}
High \emph{accuracy} is obtained for both $S^3$ and $A^3$ method, as shown in Table \ref{tabAccu}. While training time is much less for $S^3$, the inference time in root-cause analysis for $S^3$ is much more than that of $A^3$, as $S^3$ depends on sequential searching. Note, the classification formulation in $A^3$ aims to achieve the exact set of anomalous nodes. On the other hand, the $S^3$ method is an approximate method that sequentially identifies anomalous patterns and hence, the stopping criteria would be critical. The observation that the performance of $S^3$ is quite comparable to that of $A^3$ suggests a reasonable choice of the stopping criteria.

\subsection{Anomaly in node(s)}%\vspace{-5pt}
\label{subsecAnoNod}

\textbf{Dataset:} Anomaly in node(s) occurs when one node or multiple nodes fail in the system. VAR process is applied to define a graphical model for generating the nominal data. Anomaly data are simulated by introducing time delay in a specific node. The time delay will break most of the causality to and from this node (except possibly the self loop, i.e., AP of the failed node). The generated dataset is denoted as \texttt{dataset2}. For scalability analysis, a 30-node system is defined.

\textbf{Performance Evaluation:} This work is aimed at discovering failed patterns instead of recovering underlying graph. The discovered anomalous patterns can then be used for diagnosing the fault node. For instance, a failed pattern $N_{i}\to N_{j}$ discovered by root-cause analysis can be caused by the fault node $i$ or $j$. However, if multiple failed patterns are related to the node $i$, then this node can be deemed anomalous. In this regard, it is important to learn the impact of one pattern on a detected anomaly compared to another. This can facilitate a ranking of the failed patterns and enable a robust isolation of an anomalous node.

For comparison, we use VAR-based graph recovery method that is widely applied in economics and other sciences, and efficient in discovering Granger causality \cite{goebel2003investigating}. Note, the test dataset itself is synthetically generated using a VAR model with a specific time delay. Hence, the causality in such a multivariate time series is supposed to be well captured by VAR-based method.

With the given time series, a VAR model (i.e., the coefficients $A_{i,j}$) can be learnt using standard algorithm~\cite{goebel2003investigating}. The differences in coefficients  between the nominal and anomalous models are subsequently used to find out the root causes. The pattern is deemed to have failed when $\delta A_{i,j}>0.4\cdot \max\{\delta A_{i,j}\}$ where $\delta A_{i,j}=|A_{i,j}^{ano}-A_{i,j}^{nom}|$.
\begin{figure}[htbp]
  \centering
  \subfigure[Fault in Node 2]{\includegraphics[width=0.4\textwidth]{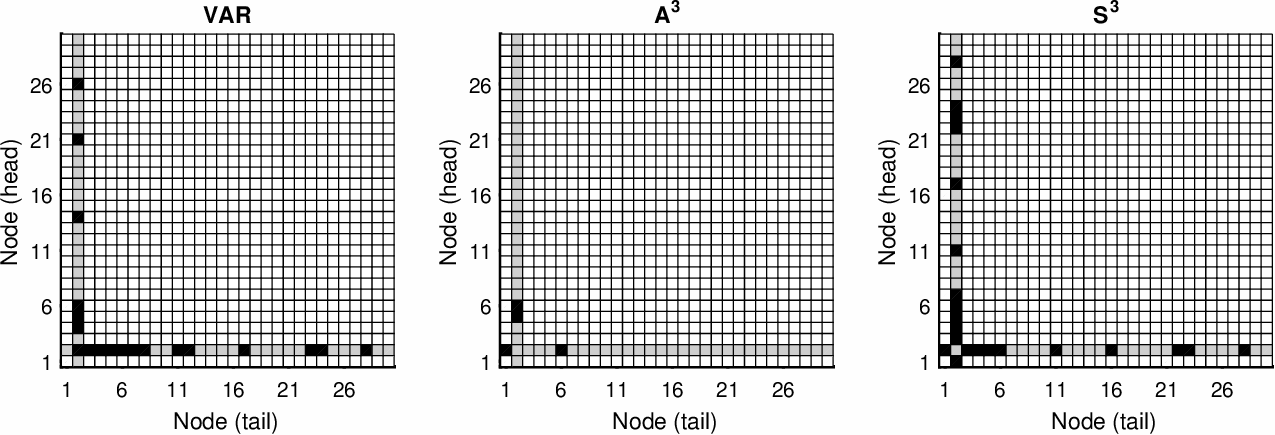}} \\ \vspace{-5pt}%\hspace{20pt} %\\
  \subfigure[Fault in Node 5]{\includegraphics[width=0.4\textwidth]{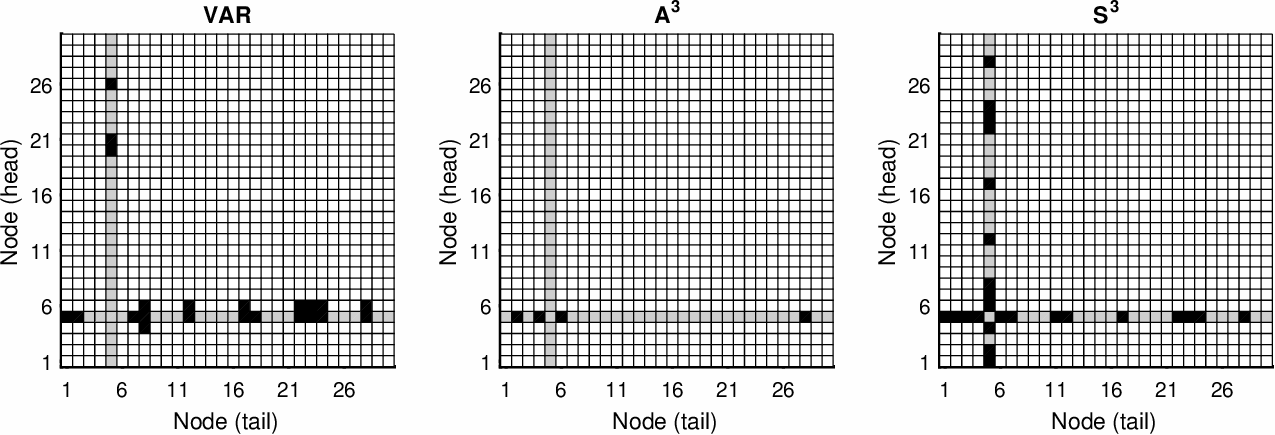}} %\\ \vspace{-5pt}
  \caption{Comparisons of VAR, $A^3$, and $S^3$ methods using \texttt{dataset2}. The boxes represent the patterns from the tail node (shown in $x$-axis) to the head node (shown in $y$-axis), and the results of root-cause analysis are in black. The gray boxes are corresponding to the simulated fault node.  } %\vspace{-5pt}
  \label{figRCA30node}
\end{figure}

The results of $A^3$, $S^3$ and VAR using \texttt{dataset2} are shown in Fig.~\ref{figRCA30node}. In panel (a), all of the changed patterns discovered by VAR, $A^3$ and $S^3$ can be attributed to node 1 (shown by the black boxes in column and row 1). Therefore, node 1 is considered as faulty by the $S^3$ method. In panel (b), VAR incorrectly discovers a significant change in the patterns in rows $4$ and $6$ but not the patterns originating from $N_5$ (We note these as errors), while $A^3$ and $S^3$ can correctly interpret the fault node $N_5$. In general, although $A^3$, $S^3$ and VAR can discover the fault node, VAR produces more false alarms. 
\begin{table}[htbp]%\vspace{-10pt}
\caption{Comparison with $A^3$, $S^3$ and VAR with dataset 2 (30 nodes).}%\vspace{3pt}
\centering
\label{tabAccu2}
\begin{tabular}{c  c c c c c c c}
\hline
  Approach & $|\{\Lambda^{ano}\}|$ &$|\{\Lambda^{\epsilon}\}|$ &$\epsilon$ (\%) \\
  \hline
  VAR   &521 &113 &21.7\\
  $A^3$   &105 &1 &0.95\\
  $S^3$  &653 &0 &0\\
\hline
\end{tabular}
\end{table}

In real applications, when more anomalous patterns are discovered incorrectly, more effort will be needed to analyze the failed patterns closely and determine the root-cause node. This will lead to more financial expenditures and time investment in finding the failed node. With this motivation, an error metric $\epsilon$ is defined by computing the ratio of incorrectly discovered anomalous patterns $|\{\Lambda^{\epsilon}\}|$ to all discovered anomalous patterns $|\{\Lambda^{ano}\}|$, i.e., $\epsilon=\frac{|\{\Lambda^{\epsilon}\}|}{|\{\Lambda^{ano}\}|}$. The results using \texttt{dataset2} are listed in Table~\ref{tabAccu2}.

The approach in \cite{qiu2012granger} can give the anomalous node in the graphical model. If we do node-based root cause analysis, then the results can be compared. For now, we only have pattern based root cause analysis. Our approaches are promising in diagnosing fault node, as shown in Fig.~\ref{figRCA30node}. We may get better results in finding out the fault node than \cite{qiu2012granger}. Compared to~\cite{zhang2014dynamic,yuan2014root}, we have an automatic algorithm to discover the root cause.

While it should be noted that the error ratio for $A^3$ and $S^3$ methods is much lower than that for VAR (i.e., lower false alarm). $A^3$ and $S^3$ methods are both \emph{scalable} as well as demonstrates better accuracy. For comparisons between $A^3$, $S^3$ and VAR, only one nominal mode is considered in Table~\ref{tabAccu2} as VAR is not directly applicable in cases with multiple nominal modes. $A^3$ and $S^3$ methods can handle multiple nominal modes and the approach has been validated in Section \ref{subsecAnoPat}.

\subsection{Validation on a Real System - TEP}
\label{subsecRealSys}
TEP data is based on a realistic simulation program of a chemical plant from the Eastman Chemical Company, USA, and it has been widely used for process monitoring community as a source of data for comparing various approaches, and a benchmark for control and monitoring studies~\cite{russell2000data,yin2012comparison}. The process consists of five major units: reactor, condenser, compressor, separator, and stripper, with 53 variables simulated including 41 measured and 12 manipulated (the agitation speed is not included in TEP dataset as it is not manipulated). 21 faults are simulated in TEP program, as the root causes of these faults are intuitively shown in \cite{russell2000data}. This work applies a subset of measurements (5 variables-- 22, 32, 48, 49, 51 treated as nodes 1 through 5 respectively) to validate the proposed approaches, and 2 faults (fault 4 and 11) are tested. Note, both faults 4 an 11 involves variable 51 (or node 5 in the present analysis) according to the ground truth.
\begin{figure}[htbp]
  \centering
  \subfigure[\scriptsize{Fault 4-$S^{3}$}]{\includegraphics[width=0.1\textwidth]{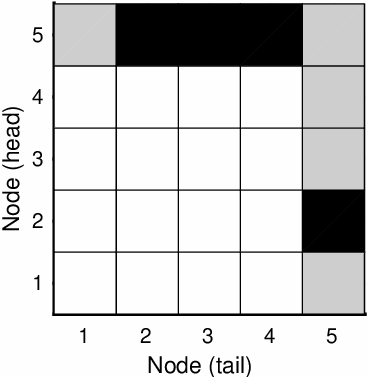}} \hspace{1pt} %\\
  \subfigure[\scriptsize{Fault 11-$S^{3}$}]{\includegraphics[width=0.1\textwidth]{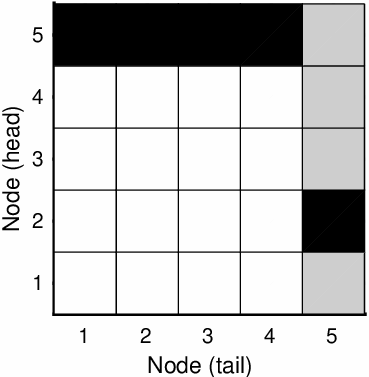}} \hspace{1pt}%\\
  \subfigure[\scriptsize{Fault 4-$A^{3}$}]{\includegraphics[width=0.11\textwidth]{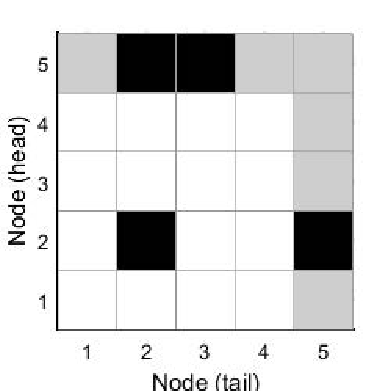}} \hspace{1pt} %\\
  \subfigure[\scriptsize{Fault 11-$A^{3}$}]{\includegraphics[width=0.11\textwidth]{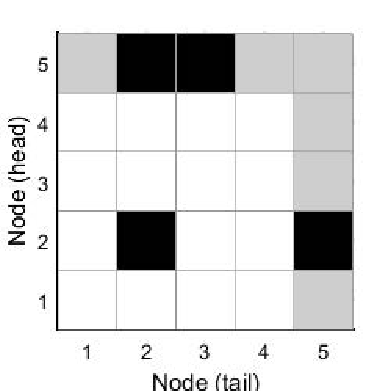}}  %\hspace{20pt}%\\
  \caption{Root-cause analysis results using $A^{3}$ and $S^{3}$ with TEP dataset. The boxes represent the patterns from the tail node (shown in $x$-axis) to the head node (shown in $y$-axis), and the results of root-cause analysis are in black. The gray boxes are corresponding to the fault node (representing the variable that most closely related to the fault, shown in \cite{russell2000data}, e.g., panel the fault occurs at Node $5$ in panel (a), the patterns that include Node $5$ as the head or the tail are and the potential fault patterns and marked as grey). } \vspace{-10pt}
  \label{figRCAtep5}
\end{figure}

Using $A^{3}$ and $S^{3}$, root-cause analysis results are shown in Fig.~\ref{figRCAtep5}. In panel (a), Node 5 is the variable most closely related to the fault, and patterns $N_{5}\to N_{2}$, $N_{2},N_{3},N_{4}\to N_{5}$ are identified as fault by $S^{3}$. The identified fault patterns will result in root cause of Fault Node $5$, as it is the shared Node in each pattern. Similarly, variable $5$ is the fault node in panel (b). In panel (c) and (d), the results are similar, while there is one error pattern in each case.

The results of the real dataset show that both $A^{3}$ and $S^{3}$ are capable of finding the root-cause of the patterns that can be reasonably interpreted to obtain the fault variable.

\section{Conclusions}
\label{secConclusion}
Based on spatiotemporal causal graphical modeling, this work presents two approaches--the sequential state switching ($S^3$) and artificial anomaly association ($A^3$)--for root-cause analysis in distributed CPSs. While $S^3$ approaches the problem in a sub-optimal sequential manner, $A^3$ takes all patterns jointly. However, experimental evidence presented here suggests that $S^3$ performs slightly better than $A^3$. With synthetic data and real data, the proposed approaches are validated and showed high accuracy in finding failed patterns and diagnose for the anomalous node. Advantages of the proposed methods include -
\begin{enumerate}
\item \emph{Ability to handle multiple nominal modes:} The STPN+ RBM framework is capable of learning multiple modes as nominal, which corresponds to diverse operation modes in most physical systems.

\item \emph{Accuracy:} The proposed approaches--$S^3$ and $A^3$-- demonstrate high accuracy in root-causes analysis. 

\item \emph{Scalability:} The approaches are scalable with the size of the system.

\item \emph{Robustness:} Compared with VAR model, the proposed approach can more effectively isolate the fault node with less incorrectly discovered patterns. %$S^3$ can also find out the maximal set of failed patterns while avoiding false positives.
\end{enumerate}

Future work will pursue: (i) inference approach in node failure including single node and multiple nodes, (ii) detection and root-cause analysis of simultaneous multiple faults in distributed complex systems.

\section*{Acknowledgement}
This work was supported by the National Science Foundation under Grant No. CNS-1464279.

\bibliographystyle{abbrv}
\bibliography{ICCPS_rca}

\end{document}